\documentclass{article}
\usepackage{spconf,amsmath,graphicx}

\usepackage{amssymb}
\usepackage[ruled,linesnumbered]{algorithm2e}
\usepackage{booktabs}
\usepackage{float}
\usepackage{bigstrut}
\usepackage{multirow}
\usepackage{balance}
\usepackage[hidelinks=true,bookmarks=false]{hyperref}
\usepackage{lineno}
\usepackage{amsopn}
\usepackage{comment}
\usepackage{color}
\usepackage{algorithmic}
\usepackage{tabularx}
\usepackage{mathrsfs}
\usepackage{makecell}
\usepackage{url,subfigure,tabulary}
\usepackage{color}
\usepackage{algorithmic}
\usepackage{makecell}

\newcommand{\ie}{\textit{i.e.}}
\newcommand{\eg}{\textit{e.g.}}

\newcommand{\etal}{\textit{et al.}}

\newcolumntype{L}[1]{>{\raggedright\arraybackslash}p{#1}}
\newcolumntype{C}[1]{>{\centering\arraybackslash}p{#1}}
\newcolumntype{R}[1]{>{\raggedleft\arraybackslash}p{#1}}

\usepackage{pifont}

\title{SCV-Stereo: Learning Stereo Matching from A Sparse Cost Volume}

\name{Hengli Wang$^{\star}$ \qquad Rui Fan$^{\dagger}$ \qquad Ming Liu$^{\star}$
}
  
\address{$^{\star}$ Hong Kong Unviersity of Science and Technology, Hong Kong SAR, China\\
      $^{\dagger}$ Tongji University, Shanghai 201804, China\\
      \normalsize\texttt{hwangdf@connect.ust.hk, rui.fan@ieee.org, eelium@ust.hk}}

\begin{document}

\maketitle

\begin{abstract}
  Convolutional neural network (CNN)-based stereo matching approaches generally require a dense cost volume (DCV) for disparity estimation. However, generating such cost volumes is computationally-intensive and memory-consuming, hindering CNN training and inference efficiency. To address this problem, we propose SCV-Stereo, a novel CNN architecture, capable of learning dense stereo matching from sparse cost volume (SCV) representations. Our inspiration is derived from the fact that DCV representations are somewhat redundant and can be replaced with SCV representations. Benefiting from these SCV representations, our SCV-Stereo can update disparity estimations in an iterative fashion for accurate and efficient stereo matching. Extensive experiments carried out on the KITTI Stereo benchmarks demonstrate that our SCV-Stereo can significantly minimize the trade-off between accuracy and efficiency for stereo matching. Our project page is \url{https://sites.google.com/view/scv-stereo}.
\end{abstract}


\section{Introduction}
\label{sec.introduction}
Stereo matching aims at finding correspondences between a pair of well-rectified left and right images \cite{fan2018road}. As a fundamental computer vision and robotics task \cite{wang2019self,wang2020applying,fan2020sne,wang2021dynamic,fan2021learning}, it has been studied extensively for decades \cite{fan2020computer}. Traditional approaches \cite{seki2017sgm,fan2018road} consist of four main steps: (i) cost computation, (ii) cost aggregation, (iii) disparity optimization, and (iv) disparity refinement \cite{fan2020computer}. With recent advances in deep learning, many data-driven approaches \cite{mayer2016large,wang2021pvstereo,psmnet,ganet} based on convolutional neural networks (CNNs) have been proposed for stereo matching. These approaches generally adopt a three-stage pipeline: (i) feature extraction, (ii) cost volume computation, and (iii) disparity estimation. Extensive studies have demonstrated their compelling performance as well as the tremendous potential to be a practical solution to 3D geometry reconstruction \cite{mayer2016large,wang2021pvstereo,psmnet,ganet}.

\begin{figure}[t]
  \centering
  \includegraphics[width=0.99\linewidth]{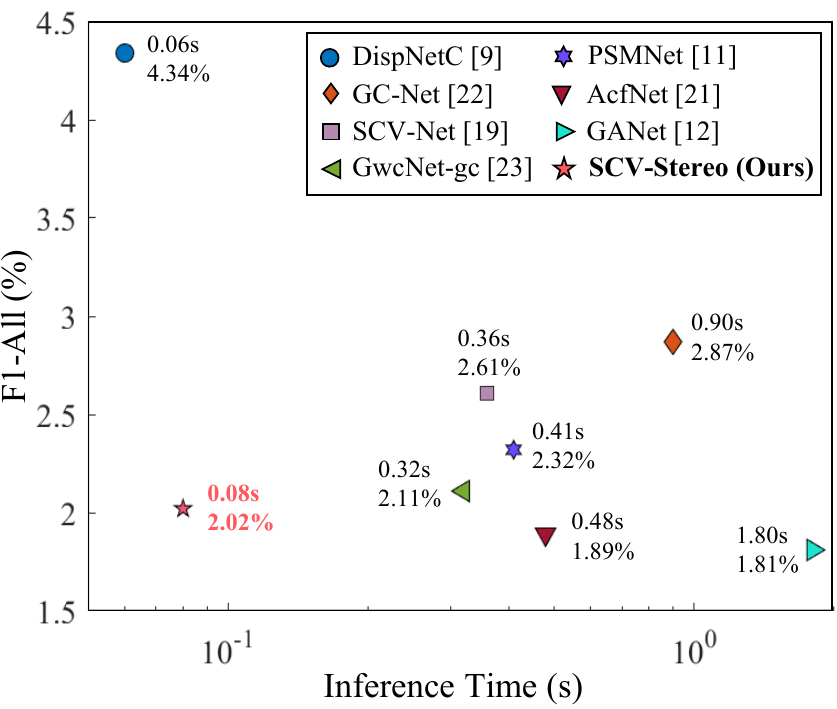}
  \caption{Evaluation results on the KITTI Stereo 2015 benchmark \cite{kitti15}, where ``F1-All'' denotes the percentage of erroneous pixels over all regions. Our SCV-Stereo can greatly minimize the trade-off between accuracy and efficiency for stereo matching.}
  \label{fig.time_error}
\end{figure}

\begin{figure*}[t]
  \centering
  \includegraphics[width=0.99\textwidth]{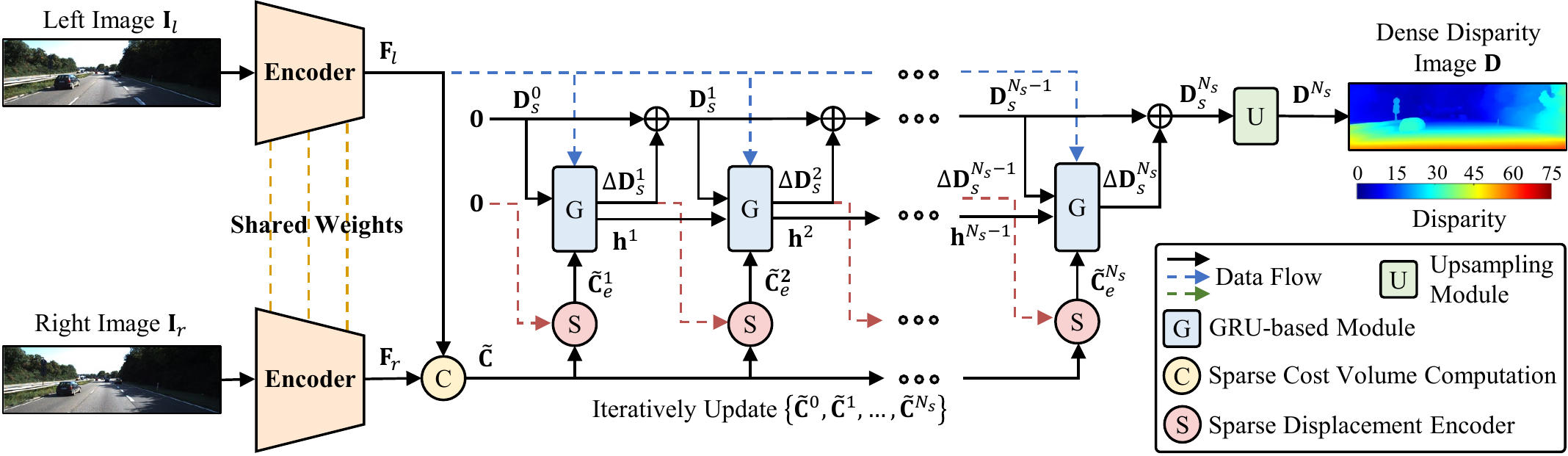}
  \caption{An overview of our SCV-Stereo. It first extracts useful visual features to build a sparse cost volume, and then employs a novel sparse displacement encoder as well as a GRU-based module to iteratively update dense disparity estimations.}
  \label{fig.framework}
\end{figure*}

According to the manner of cost volume computation, existing data-driven stereo matching approaches can be grouped into two classes: (i) 2D CNN-based \cite{mayer2016large,wang2021pvstereo} and (ii) 3D CNN-based \cite{psmnet,ganet}. The former generally adopt a correlation layer, \eg, using dot-product operation, to produce cost volumes. In contrast, the latter first perform left and right feature concatenation and then employ 3D convolution layers to generate cost volumes. While 3D CNN-based approaches perform more accurately than 2D CNN-based approaches, they generally present poor performance in memory consumption and inference speed, making them inapplicable in practice \cite{ganet}. Hence, boosting the accuracy of 2D CNN-based approaches while maintaining their computational efficiency has become a promising research direction.

Recently, Teed \etal \cite{teed2020raft} proposed RAFT, a novel 2D CNN-based architecture, for optical flow estimation. Unlike other approaches for dense correspondence estimation, RAFT first constructs an all-pairs dense cost volume (DCV) by computing the pair-wise dot product between the left and right feature maps \cite{teed2020raft}. Then, it iteratively updates the optical flow estimation at a single resolution with a gated recurrent unit (GRU) \cite{cho2014learning}. Our previous work has employed this effective architecture for stereo matching \cite{wang2021pvstereo}. However, the all-pairs DCV can consume a lot of memory and computational resources, limiting the feature maps used to only \textbf{1/8} image resolution \cite{teed2020raft}. This further frustrates the performance of dense correspondence estimation.

To address this problem, we propose SCV-Stereo, a novel 2D CNN-based stereo matching network, in this paper. It can greatly minimize the trade-off between accuracy and efficiency for stereo matching, as shown in Fig.~\ref{fig.time_error}. Our SCV-Stereo adopts a sparse cost volume (SCV) representation learning scheme, as illustrated in Fig.~\ref{fig.framework}. An SCV representation only stores the best $K$ matching costs for each pixel to avoid the intensive computation and memory consumption of existing DCV representations. We also employ a sparse displacement encoder to extract useful visual information from SCV representations. Using such sparse representations, our SCV-Stereo can iteratively update disparity estimations at \textbf{1/4} image resolution for accurate and efficient stereo matching.

\section{Methodology}
\label{sec.mechodology}

\subsection{Network Architecture}
\label{sec.network_architecture}
SCV-Stereo generates a dense disparity image $\mathbf{D}$ by comparing the difference between a pair of well-rectified left image $\mathbf{I}_{l}$ and right image $\mathbf{I}_{r}$, as illustrated in Fig.~\ref{fig.framework}. It consists of three stages: (i) feature extraction, (ii) sparse cost volume computation, and (iii) iterative disparity update. We also employ a sparse displacement encoder to provide useful information for disparity estimation. The details are introduced as follows.

\subsubsection{Feature Extraction}
We first use two weight-sharing encoders to separately extract useful visual features $\mathbf{F}_{l}$ and $\mathbf{F}_{r}$ from $\mathbf{I}_{l}$ and $\mathbf{I}_{r}$. Each encoder consists of several residual blocks \cite{he2016deep}. $\mathbf{F}_{l}$ and $\mathbf{F}_{r}$ have the size of $H \times W \times C$ at 1/4 image resolution of $\mathbf{I}_{l}$ and $\mathbf{I}_{r}$.

\subsubsection{Sparse Cost Volume Computation}
Existing 2D approaches often compute a DCV, denoted as $\mathbf{C}$, to model the similarity for possible matching pairs between $\mathbf{F}_{l}$ and $\mathbf{F}_{r}$ \cite{teed2020raft}, which can be formulated as follows:
\begin{equation}
  \mathbf{C}(\mathbf{p}, \mathbf{d}) = \langle \mathbf{F}_{l}(\mathbf{p}), \mathbf{F}_{r}(\mathbf{p} - \mathbf{d}) \rangle,
\end{equation}
where $\mathcal{P}$ denotes the domain of $\mathbf{F}_{l}$ and $\mathbf{p} \in \mathcal{P}$ denotes the pixel coordinates; $\mathcal{D}$ denotes the disparity candidate set and $\mathbf{d} \in \mathcal{D}$ denotes the candidates; and $\langle \cdot,\cdot \rangle$ denotes the dot product operation.

Referring to \cite{fan2018road}, it is unnecessary to have DCV representations for disparity estimation, and an SCV representation filling with inconsecutive and discrete matching costs is feasible to produce accurate disparity images. Therefore, we follow \cite{jiang2021learning} and use an SCV representation $\widetilde{\mathbf{C}}$ that only stores the best $K$ matching costs for each pixel:
\begin{equation}
  \begin{aligned}
    \widetilde{\mathbf{C}}       & = \left\{ \mathbf{C}(\mathbf{p}, \mathbf{d}) \mid \mathbf{p} \in \mathcal{P}, \mathbf{d} \in \mathcal{D}_{\mathbf{p}}^{K} \right\},                                                \\
    \mathcal{D}_{\mathbf{p}}^{K} & = \underset{\widetilde{\mathcal{D}} \subset \mathcal{D},|\widetilde{\mathcal{D}}|=K}{\arg \max } \sum_{\mathbf{d} \in \widetilde{\mathcal{D}}} \mathbf{C}(\mathbf{p}, \mathbf{d}).
  \end{aligned}
  \label{eq.sparse}
\end{equation}
We achieve this by using a $k$-nearest neighbors ($k$NN) module \cite{johnson2019billion}. Please note that only the gradients correlated to the selected $K$ matching costs are back-propagated for CNN parameter optimization. Considering the trade-off between accuracy and efficiency, we set $K = 8$, and it is much smaller than the number of disparity candidates in $\mathcal{D}$. Moreover, after constructing $\widetilde{\mathbf{C}}$, we only need to update the cost volume coordinates instead of re-computing dot products as usual. Such a unique SCV representation enables our SCV-Stereo to iteratively update disparity estimations at 1/4 image resolution for accurate and efficient stereo matching.

\subsubsection{Sparse Displacement Encoder}
Inspired by RAFT \cite{teed2020raft}, we adopt an iterative disparity update scheme for disparity refinement, as shown in Fig.~\ref{fig.framework}. Specifically, at step $i$, we first estimate the residual disparity $\Delta \mathbf{D}^{i}_{s}$. The disparity estimation $\mathbf{D}^{i}_{s}$ is then updated by $\mathbf{D}^{i}_{s} = \mathbf{D}^{i-1}_{s} + \Delta \mathbf{D}^{i}_{s}$. To perform this iterative disparity update scheme based on the constructed SCV representation $\widetilde{\mathbf{C}}$, we design a sparse displacement encoder, which contains a displacement update phase and a multi-scale encoding phase.

During the displacement update phase, we update $\widetilde{\mathbf{C}}$ based on the current disparity residual for further residual estimation, \ie, $\widetilde{\mathbf{C}}^{i}(\mathbf{p},\mathbf{d}^{i-1}-\Delta \mathbf{D}^{i-1}_{s}) = \widetilde{\mathbf{C}}^{i-1}(\mathbf{p},\mathbf{d}^{i-1})$. During the multi-scale encoding phase, we aim to encode $\widetilde{\mathbf{C}}^{i}$ to a dense tensor for further iterative disparity update. We first construct a five-level SCV pyramid $\{\widetilde{\mathbf{C}}^{i}_{l}\}_{l=0,1,2,3,4}$, where $\mathbf{d}^{i}_{l} = \mathbf{d}^{i} / 2^{l}$ for $\widetilde{\mathbf{C}}^{i}_{l}(\mathbf{p},\mathbf{d}^{i}_{l})$. Since the coordinates $\mathbf{d}^{i}_{l}$ can be float numbers, we then follow \cite{jiang2021learning} and linearly propagate the matching costs to the two nearest coordinates. Afterwards, we can build a 1D dense tensor (with the size of $(2d_{\max}+1)$) for each pixel $\mathbf{p} \in \mathcal{P}$ in $\widetilde{\mathbf{C}}^{i}_{l}$ as follows:
\begin{equation}
  \left\{ \left(\mathbf{d}^{i}_{l}, \widetilde{\mathbf{C}}^{i}_{l}(\mathbf{p},\mathbf{d}^{i}_{l})\right) \mid \left\|\mathbf{d}^{i}_{l}\right\|_{\infty} \leqslant d_{\max}, \mathbf{d}^{i} \in \mathcal{D}_{\mathbf{p}}^{K} \right\},
  \label{eq.sparse_to_dense}
\end{equation}
where $\left\| \cdot \right\|_{\infty}$ denotes the infinity norm; and the missing matching costs are filled by 0. Based on \eqref{eq.sparse_to_dense}, $\{\widetilde{\mathbf{C}}^{i}_{l}\}_{l=0,1,2,3,4}$ can be transformed to dense tensors, which are then reshaped and concatenated to form a 3D dense tensor $\widetilde{\mathbf{C}}^{i}_{e}$. $\widetilde{\mathbf{C}}^{i}_{e}$ can provide useful position information and matching costs for further iterative disparity update.

\begin{table}[t]
  \centering
  \caption{Evaluation results of our SCV-Stereo with different setups on the Scene Flow dataset \cite{mayer2016large}. The adopted setup (the best result) is shown in bold type.}
  \begin{tabular}{C{0.6cm}C{1.9cm}C{1.8cm}C{0.6cm}C{1.6cm}}
    \toprule
    No. & Cost Volume & Resolution & $K$ & AEPE~(px)     \\ \midrule
    (a) & Dense       & 1/8        & --  & 1.05          \\
    (b) & Sparse      & 1/8        & 8   & 1.28          \\ \midrule
    (c) & Sparse      & 1/4        & 1   & 2.36          \\
    (d) & Sparse      & 1/4        & 2   & 1.52          \\
    (e) & Sparse      & 1/4        & 4   & 1.01          \\ \midrule
    (f) & Sparse      & 1/4        & 8   & \textbf{0.93} \\
    \bottomrule
  \end{tabular}
  \label{tab.ablation}
\end{table}

\subsubsection{Iterative Disparity Update}
As mentioned above, we follow the paradigm of RAFT \cite{teed2020raft} and utilize a GRU-based module \cite{cho2014learning} to iteratively update disparity estimations $\{ \mathbf{D}^{1}_{s}, \dots, \mathbf{D}^{N_s}_{s} \}$ at 1/4 image resolution (with an initialization $\mathbf{D}^0_s = 0$), as shown in Fig.~\ref{fig.framework}. Specifically, in step $i$, we denote the concatenation of $\mathbf{D}^{i-1}_{s}$, $\widetilde{\mathbf{C}}^{i}_{e}$ and $\mathbf{F}_{l}$ as $\mathbf{x}^{i}$, and then send it to the GRU-based module with the hidden state $\mathbf{h}^{i}$ to estimate the residual disparity $\Delta \mathbf{D}^{i}_{s}$, which then updates the disparity estimation $\mathbf{D}^{i}_{s}$ by $\mathbf{D}^{i}_{s} = \mathbf{D}^{i-1}_{s} + \Delta \mathbf{D}^{i}_{s}$. We also use an upsampling module that consists of an upsampling layer followed by two convolution layers to provide disparity estimations at full image resolution $\{ \mathbf{D}^{1}, \dots, \mathbf{D}^{N_s} \}$.

\subsection{Loss Function}
\label{sec.loss_function}
Our adopted loss function $\mathcal{L}$ is defined as follows:
\begin{align}
  \mathcal{L} = \frac{1}{\sum_{i=1}^{N_s} \alpha^{N_s-i}} & \sum_{i=1}^{N_s} \frac{\alpha^{N_s-i}}{N_{\widetilde{\mathbf{D}}}} \sum_{\mathbf{p}\in\widetilde{\mathbf{D}}} l(|\widetilde{\mathbf{D}}(\mathbf{p})-\mathbf{D}^{i}(\mathbf{p})|), \notag \\
  l(x)     =                                              & \left\{\begin{array}{ll}
    {x-0.5,}     & {x \geq 1} \\
    {x^{2} / 2,} & {x<1}
  \end{array},\right.
  \label{eq.guiding_loss}
\end{align}
where $l(\cdot)$ denotes the smooth L1 loss; $\widetilde{\mathbf{D}}$ denotes ground-truth labels; $N_{\widetilde{\mathbf{D}}}$ is the number of valid pixels in $\widetilde{\mathbf{D}}$; and we set $\alpha = 0.8$ in our experiments, as suggested in \cite{teed2020raft}.

Although SCV-Net \cite{lu2018sparse} was also developed based on SCV representations, there exist major differences between it and our approach. Firstly, we present a novel SCV representation and adopt an iterative disparity update scheme, which is different from SCV-Net \cite{lu2018sparse}. Moreover, the novel SCV representation and network architecture enable our SCV-Stereo to greatly outperform SCV-Net \cite{lu2018sparse} in terms of both accuracy and speed, as shown in Fig.~\ref{fig.time_error} and Table~\ref{tab.disparity}.

\begin{figure*}[t]
  \centering
  \includegraphics[width=0.99\textwidth]{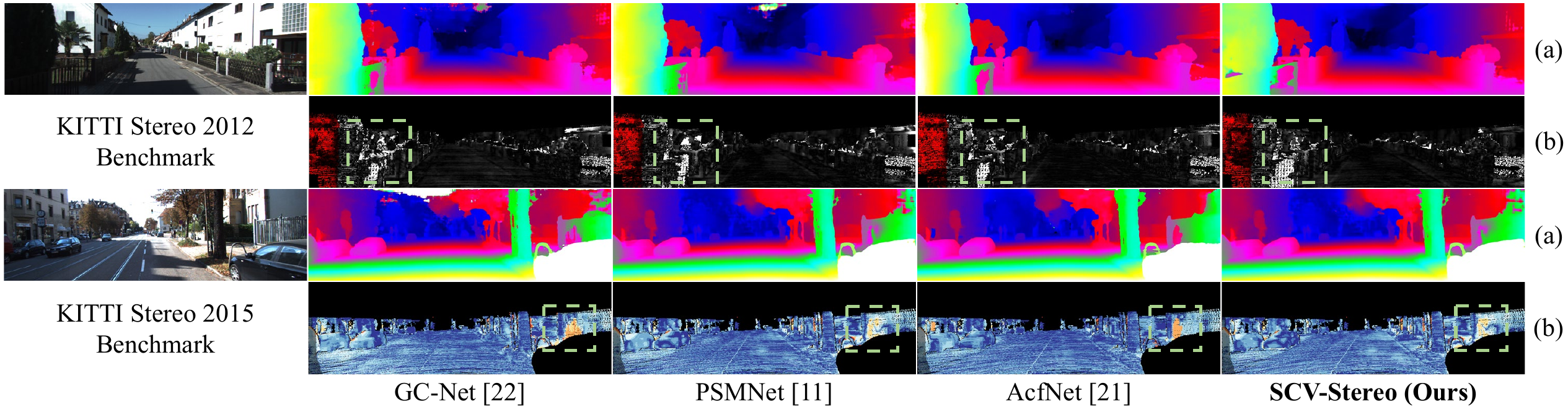}
  \caption{Examples on the KITTI Stereo benchmarks \cite{kitti12,kitti15}, where rows (a) and (b) show the disparity estimations and the corresponding error maps, respectively. Significantly improved regions are highlighted with green dashed boxes.}
  \label{fig.benchmark}
\end{figure*}

\section{Experimental Results}
\label{sec.experimental_results}

\subsection{Datasets and Implementation Details}
\label{sec.datasets_and_implementation_details}
In our experiments, we use three datasets, the Scene Flow dataset \cite{mayer2016large}, the KITTI Stereo 2012 \cite{kitti12} and Stereo 2015 \cite{kitti15} datasets, to demonstrate the effectiveness of our approach. The Scene Flow dataset \cite{mayer2016large} was rendered from synthetic scenarios. Differently, the KITTI stereo datasets \cite{kitti12,kitti15} were created based on real-world driving scenarios. Moreover, the KITTI stereo datasets also provide public benchmarks for fair performance comparison.

For the implementation, we set $d_{\max} = 4$ and $N_s = 8$. We use PyTorch to implement our model, and train it using the Adam optimizer. The training scheme follows \cite{ganet,zhang2020adaptive} for fair performance comparison. Afterwards, we use $\mathbf{D}^{N_s}$ as the dense disparity prediction $\mathbf{D}$ to evaluate the model's performance. Moreover, we utilize two standard evaluation metrics, (i) the average end-point error (AEPE) that measures the difference between the disparity estimations and ground-truth labels and (ii) the percentage of the bad pixels whose error is larger than 3 pixels (F1) \cite{kitti12,kitti15}.

We first conduct ablation studies on the Scene Flow dataset \cite{mayer2016large} to validate the effectiveness of our proposed architecture, as presented in Section~\ref{sec.ablation_study}. Then, we submit the results of our SCV-Stereo to the two KITTI Stereo benchmarks \cite{kitti12,kitti15}, as shown in Section~\ref{sec.evaluations_on_the_public_benchmarks}.

\begin{table}[t]
  \small
  \centering
  \caption{Evaluation results on the KITTI Stereo 2012$^{1}$ \cite{kitti12}  and Stereo 2015$^{2}$ \cite{kitti15} benchmarks. ``Noc'' and ``All'' represent the F1~($\%$) for non-occluded pixels and all pixels, respectively \cite{kitti12,kitti15}. Best results are shown in bold type, and the red number indicates the ranking among these eight approaches.}
  \begin{tabular}{L{2.55cm}C{0.7cm}C{0.7cm}C{0.7cm}C{0.7cm}C{0.7cm}}
    \toprule
    \multicolumn{1}{l}{\multirow{2}{*}{Approach}} & \multicolumn{2}{c}{KITTI 2012} & \multicolumn{2}{c}{KITTI 2015} & \multicolumn{1}{c}{\multirow{2}{*}{\makecell{Time                                                               \\(s)}}}                                                               \\ \cmidrule(l){2-3} \cmidrule(l){4-5}
    \multicolumn{1}{c}{}                          & Noc                            & All                            & Noc                                               & All                                                         \\ \midrule
    DispNetC \cite{mayer2016large}                & 4.11$\color{red}^7$            & 4.65$\color{red}^7$            & 4.05$\color{red}^{8}$                            & 4.34$\color{red}^{8}$       & \textbf{0.06}$\color{red}^1$ \\
    GC-Net \cite{kendall2017end}                  & 1.77$\color{red}^6$            & 2.30$\color{red}^6$            & 2.61$\color{red}^7$                               & 2.87$\color{red}^7$          & 0.90$\color{red}^7$          \\
    SCV-Net \cite{lu2018sparse}                   & --                             & --                             & 2.41$\color{red}^6$                               & 2.61$\color{red}^6$          & 0.36$\color{red}^4$          \\
    PSMNet \cite{psmnet}                          & 1.49$\color{red}^5$            & 1.89$\color{red}^5$            & 2.14$\color{red}^5$                               & 2.32$\color{red}^5$          & 0.41$\color{red}^5$          \\
    GwcNet-gc \cite{gwcnet}                       & 1.32$\color{red}^4$            & 1.70$\color{red}^4$            & 1.92$\color{red}^4$                               & 2.11$\color{red}^4$          & 0.32$\color{red}^3$          \\
    AcfNet \cite{zhang2020adaptive}               & \textbf{1.17}$\color{red}^1$            & \textbf{1.54}$\color{red}^1$            & 1.72$\color{red}^2$                               & 1.89$\color{red}^2$          & 0.48$\color{red}^6$          \\
    GANet \cite{ganet}                            & 1.19$\color{red}^2$            & 1.60$\color{red}^2$            & \textbf{1.63}$\color{red}^1$                               & \textbf{1.81}$\color{red}^1$          & 1.80$\color{red}^{8}$       \\ \midrule
    \textbf{SCV-Stereo (Ours)}                    & 1.27$\color{red}^3$            & 1.68$\color{red}^3$            & 1.84$\color{red}^3$                               & 2.02$\color{red}^3$          & 0.08$\color{red}^2$          \\ \bottomrule
  \end{tabular}
  \label{tab.disparity}
\end{table}

\footnotetext[1]{\url{http://cvlibs.net/datasets/kitti/eval_stereo_flow.php?benchmark=stereo
  }}
\footnotetext[2]{\url{http://cvlibs.net/datasets/kitti/eval_scene_flow.php?benchmark=stereo}}

\subsection{Ablation Study}
\label{sec.ablation_study}
Table~\ref{tab.ablation} presents the performance of our SCV-Stereo with different setups on the Scene Flow dataset \cite{mayer2016large}. By comparing (a) and (b) with (f), we can see that using the feature maps with a larger resolution can greatly improve the performance, and our SCV representation is more effective than the DCV representation for stereo matching. Please note that the DCV representation can consume significant memory, and therefore, we can only use feature maps at 1/8 resolution. Moreover, (c)--(f) show that our SCV-Stereo can even work with $K=1$, and a larger $K$ can provide a better performance. As mentioned previously, we consider the trade-off between accuracy and efficiency for stereo matching, and adopt $K = 8$ in our SCV-Stereo. All the analysis demonstrates the effectiveness of our proposed architecture.

\subsection{Evaluations on the Public Benchmarks}
\label{sec.evaluations_on_the_public_benchmarks}
Table~\ref{tab.disparity} presents the online leaderboards of the KITTI Stereo benchmarks \cite{kitti12,kitti15}, and Fig.~\ref{fig.time_error} visualizes the results on the KITTI Stereo 2015 \cite{kitti15} benchmark. It is observed that our SCV-Stereo can outperform several existing stereo matching approaches while achieving real-time performance. In addition, according to the rankings in Table~\ref{tab.disparity}, we can see that our SCV-Stereo greatly minimizes the trade-off between accuracy and efficiency for stereo matching. Although AcfNet~\cite{zhang2020adaptive} is a challenging baseline, it is exciting to see that our SCV-Stereo can achieve an almost 6$\times$ higher inference speed with only an accuracy decrease of 7$\%$ compared to AcfNet \cite{zhang2020adaptive}. Moreover, Fig.~\ref{fig.benchmark} shows qualitative results on the KITTI Stereo benchmarks, where it can be seen that our SCV-Stereo can provide more robust and accurate disparity estimations.

\section{Conclusion and Future Work}
\label{sec.conclusion_and_future_work}
This paper proposed SCV-Stereo, a novel stereo matching network that adopts a sparse cost volume representation learning scheme and an iterative disparity update scheme. Our approach can greatly reduce the computation complexity and memory consumption of existing dense cost volume representations while providing highly accurate disparity estimations. Experiments on the KITTI Stereo benchmarks show that our SCV-Stereo can greatly minimize the trade-off between accuracy and efficiency for stereo matching. We believe that our approach has laid a foundation for future stereo matching research where the cost volume memory consumption is no longer a limiting factor. The use of this sparse cost volume representation is also encouraging in other dense correspondence matching tasks, such as scene flow estimation.


\clearpage
\bibliographystyle{IEEEbib}
\bibliography{egbib}

\end{document}